\documentclass{bmvc2k}
\usepackage{float}
\usepackage{color}
\usepackage{array}
\usepackage{booktabs}
\usepackage{makecell}

\usepackage{colortbl}% http://ctan.org/pkg/colortbl
\usepackage{amsmath}% http://ctan.org/pkg/amsmath
\usepackage{xcolor}% http://ctan.org/pkg/xcolor
\colorlet{tableheadcolor}{gray!25} % Table header colour = 25% gray
\colorlet{tablerowcolor}{gray!20} % Table row separator colour = 10% gray
\newcommand{\rowcol}{\rowcolor{tablerowcolor}} %

\title{SPARC: Sparse Render-and-Compare for CAD model alignment in a single RGB Image}
\addauthor{Florian Langer}{fml35@cam.ac.uk}{1}
\addauthor{Gwangbin Bae}{gb585@cam.ac.uk}{1}
\addauthor{Ignas Budvytis}{ib255@cam.ac.uk}{1}
\addauthor{Roberto Cipolla}{rc10001@cam.ac.uk}{1}

\addinstitution{
 Department of Engineering\\
 University of Cambridge\\
 Cambridge, UK
}

\runninghead{Langer et al.}{SPARC: Sparse Render-and-Compare for CAD alignment}

\begin{document}

\maketitle
\vspace{-0.8cm}
\begin{abstract}
Estimating 3D shapes and poses of static objects from a single image has important applications for robotics, augmented reality and digital content creation. Often this is done through direct mesh predictions \cite{meshrcnn,pixel2mesh,total_3D_understanding} which produces unrealistic, overly tessellated shapes or by formulating shape prediction as a retrieval task followed by CAD model alignment \cite{mask2cad,patch2cad,geometric_correspondence_fields,roca}. Directly predicting CAD model poses from 2D image features is difficult and inaccurate \cite{mask2cad,patch2cad}. Some works, such as ROCA \cite{roca}, regress normalised object coordinates and use those for computing poses. While this can produce more accurate pose estimates, predicting normalised object coordinates is susceptible to systematic failure. Leveraging efficient transformer architectures \cite{perceiver} we demonstrate that a sparse, iterative, render-and-compare approach is more accurate and robust than relying on normalised object coordinates.
For this we combine 2D image information including sparse depth and surface normal values which we estimate directly from the image with 3D CAD model information in early fusion. In particular, we reproject points sampled from the CAD model in an initial, random pose and compute their depth and surface normal values.
This combined information is the input to a pose prediction network, SPARC-Net which we train to predict a 9 DoF CAD model pose update. The CAD model is reprojected again and the next pose update is predicted. Our alignment procedure converges after just 3 iterations, improving the state-of-the-art performance on the challenging real-world dataset ScanNet \cite{scannet} from $25.0\%$ \cite{roca} to $31.8\%$ instance alignment accuracy. Code will be released under \url{https://github.com/florianlanger/SPARC}.
\end{abstract}
\vspace{-0.2cm}
\section{Introduction}
\begin{figure*}[t]
    \centering
    \includegraphics[width=1.0\linewidth]{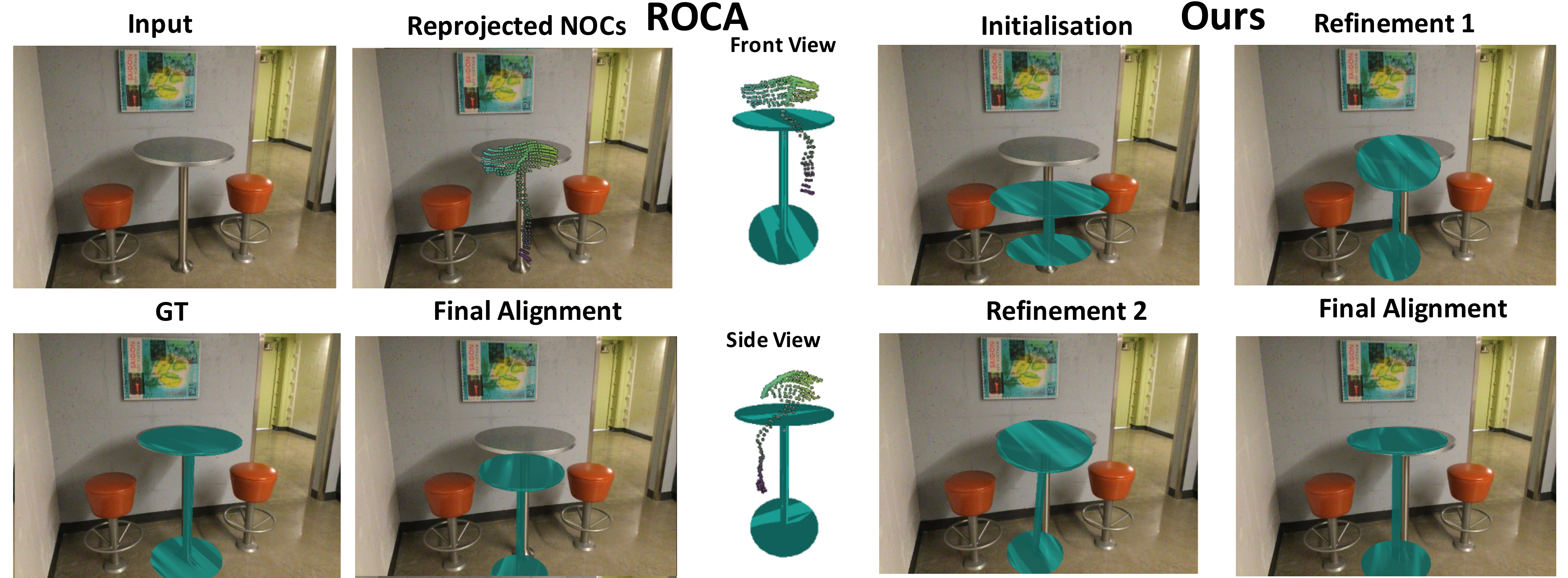}
    \vspace{-0.3cm}
    \caption{\textbf{Comparison of our CAD model alignment approach to ROCA \cite{roca}.} For each pixel of the detected object ROCA predicts the 3D normalised object coordinates (NOCs) in a canonical, normalised frame. However, those predictions are susceptible to systematic offsets (see Front View and Side View). While the reprojected NOCs match the image, the corresponding CAD model alignment does not. Our approach in contrast reprojects points and surface normals sampled from the CAD model in an initial pose into the image and uses those to predict pose updates. By iteratively updating the pose and reprojecting our system achieves precise CAD model alignments.}
    \vspace{-0.2cm}
    \label{fig_intro}
\end{figure*}
Previous work on shape and pose prediction can be classified into two different types of methods relying either on shape generation \cite{meshrcnn,pixel2mesh,total_3D_understanding} or shape retrieval \cite{mask2cad,patch2cad,geometric_correspondence_fields,roca}. Generative approaches usually struggle to produce realistic object shapes.
% or precise shapes matching the objects in the image.
For methods relying on shape retrieval one of the key challenges is to align the retrieved CAD model to the object detected in the image \cite{mask2cad,patch2cad,geometric_correspondence_fields,roca}. Many existing approaches directly regress object poses from the 2D features of the image \cite{mask2cad,patch2cad,vid2cad,engelmann2020points}. However, this produces approximate poses rather than accurate alignments. ROCA \cite{roca} follows a more geometric approach in which they predict normalised object coordinates (NOCs) \cite{vitruvian_manifold}, dense correspondences from 2D pixel to 3D points in a canonical object space, that are used to compute the object pose. The fundamental issue with learning NOCs is that it is unclear how  different shapes should be registered with each other. This means that the NOCs do not generalise well between CAD models and predicting them often fails with a systematic offset leading to a displacement in the final alignment (see Figure \ref{fig_intro}).
Rather than using NOCs we propose a sparse, render-and-compare approach.
We combine 2D image information including sparse depth and surface normals along with RGB colors with 3D CAD model information. Specifically, we initialise a CAD model in a generic pose and reproject points and surface normals sampled uniformly from its surface onto the image plane.
% For the 2D image information we estimate per-pixel depth and surface normals and concatenate these along with their pixel coordinates and RGB values to make up the second part of the network input.
This combined information is used by our pose update prediction network, SPARC-Net, to estimate a 9-DoF pose update.
After adding the predicted pose update to the initial pose we reproject the CAD model again and estimate the next pose update step. Repeating this procedure we obtain the final pose in just three iterations.\\
Having access to estimated normal and depth values from the image and reprojected normals and depth from the CAD model allows the network to evaluate the current pose and predict an accurate pose update by easily comparing observed (image) and projected (CAD model) information. This is in contrast to other approaches that do not make use of any shape information when predicting object poses 
\cite{mask2cad,patch2cad} or that rely on shape encodings
\cite{roca} which seems to be a more difficult learning task than using render-and-compare.
Note also that because of the render-and-compare our approach does not require the ability to register different CAD models with each other nor for the network to memorise 3D shapes.\\
We choose the Perceiver \cite{perceiver} architecture over traditional CNNs for the pose prediction network as Perceivers \cite{perceiver} allow for an efficient processing of the sparse object reprojection as they do not require a full, image-sized input. Also Perceivers \cite{perceiver} have linear time and memory complexity in terms of input as opposed to traditional transformers \cite{attention_all_need}. Not only do sparse inputs reduce the memory and network complexity, but we also show that they improve the alignment accuracy by avoiding overfitting.
% Further we show that using very sparse inputs improves the alignment accuracy as it avoids overfitting, while simultaneously reducing memory and network complexity by two orders of magnitude.\\
% After adding the predicted pose update to the initial pose we reproject the CAD model again and estimate the next pose update step. Repeating this procedure we obtain the final pose after just three iterations.\\
% At the heart of our approach lies the intuitive idea 
% that reprojecting a CAD model into the image plane gives the alignment network direct access to valuable information about the CAD model. This is in contrast to other approaches that do not make use of any shape information when predicting object poses 
% \cite{mask2cad,patch2cad} or that rely on shape encodings
% \cite{roca} which seems to be a more difficult learning task than using render-and-compare.
Using our approach we improve upon the state-of-the art performance on the challenging real-world dataset ScanNet \cite{scannet} from $25.0\%$ instance alignment accuracy to $31.8\%$ instance alignment accuracy.\\
\textbf{Our contributions include:}
\begin{itemize}
    \item a \textbf{novel, sparse, render-and-compare method} that achieves state-of-the-art results in CAD model alignment from a single RGB image
    \item a \textbf{demonstration of} how 3D CAD model information and 2D image information can be effectively combined with \textbf{early-fusion}
    \item \textbf{highlighting that sparse inputs improve alignment accuracy} while at the same time \textbf{significantly decreasing memory and compute complexity}.
\end{itemize}

    % \item ighlighting fundamental issues when attempting to register differently shaped CAD models with each other for estimating their normalised object coordinates.

\vspace{-0.2cm}

\section{Related Work}\vspace{-0.2cm}
In this section we discuss relevant works including shape estimation from a single image, CAD model retrieval and alignment and efficient transformer architectures.\\
\textbf{Shape estimation from a single RGB image.}
Most works on shape estimation from a single image are structured such as to use an RGB image as input to a neural network and directly predict a 3D shape using some particular representation. Various representations have been explored ranging from voxels \cite{voxels} to point clouds \cite{point_set,pointclouds}, meshes \cite{pixel2mesh,meshrcnn,TMN,total_3D_understanding}, packed spheres \cite{point_set}, binary space partitioning \cite{binary_planes}, convex polytopes \cite{polytopes}, signed distance fields \cite{sdf} to other implicit representations \cite{nerf}. However, this is a difficult learning task and regardless of the representation chosen most approaches struggle to predict realistic shapes. Because of these issues we use a retrieval-based method.\\
\textbf{CAD model retrieval and alignment.}
Rather than generating shapes a second class of approaches estimates 3D shapes by retrieving from large-scale CAD model databases \cite{chang2015shapenet}. The large number of available CAD models ensures that appropriate CAD models can be retrieved which can reconstruct the scene in a clean and compact representation that can be easily consumed for downstream applications. Existing approaches differ significantly in their procedures for aligning the retrieved CAD models to the image. Mask2CAD \cite{mask2cad} as well as \cite{patch2cad,vid2cad} simply regress the object pose from image features. However, this produces inaccurate pose estimates and requires the network to memorize every CAD model shape, consequently performing poorly for unseen CAD models. \cite{langer_leveraging_shape} explicitly leverages the geometry of the retrieved CAD model by estimating 2D-3D keypoint matches but depend on accurate segmentation masks and require exact correspondence between CAD model and the object observed in the image. ROCA \cite{roca} learns dense correspondences between 2D pixels and 3D object coordinates in a normalised, canonical frame and use those for computing the object pose. While these dense correspondences are more robust than sparse ones, they often fail systematically, leading to constant offsets in the pose alignments (see Supp. mat.).\\
% These systematic failures arise as it is fundamentally unclear how points from differently shaped CAD models should be registered with each other, such that networks struggle to accurately predict them.\\
Another class of methods uses render-and-compare to iteratively update pose estimates  \cite{geometric_correspondence_fields,Im2CAD}. These tend to be more precise, but rely on good pose initialisation \cite{geometric_correspondence_fields,Im2CAD}.
 Traditionally these have been very slow due to the time consuming rendering process and the large number of renderings required.
 The large number of renderings is needed as existing render-and-compare works for shape and pose estimation learn a comparison function which they directly minimise at test time using gradient descent which requires a large number of steps \cite{geometric_correspondence_fields,Im2CAD} (on the order of 100s to 1000). In spirit our approach is most similar to the render-and-compare approaches. However, the advantage of our method is that it does not require rendering of the full object, but just a simple reprojection of a small amount of 3D points (e.g. 100,see \ref{sec_ablation}), and it directly learns pose updates rather than a comparison function, making our approach a lot faster than \cite{Im2CAD,geometric_correspondence_fields}.\\
% \textbf{Monocular depth and surface normal estimation.} Estimating depth from a single RGB image is a challenging task as limited geometric information is available. Recent learning-based methods \cite{GB-mono-2014-eigen,GB-mono-2015-eigen,GB-mono-2015-liu,GB-mono-2016-laina,GB-mono-2017-kuznietsov,GB-mono-2019-BTS,GB-mono-2019-VNL,GB-mono-2020-adabins,GB-mono-2020-dav,GB-mono-2021-transdepth} use deep convolutional neural networks (DCNNs) to extract features and regress the per-pixel metric depth. While most methods solve depth estimation as regression, other methods recast the problem as classification \cite{GB-mono-2017-cao} or ordinal regression (i.e. classification on ordered thresholds) \cite{GB-mono-2018-DORN} by discretizing the output depth. Similar to depth estimation, surface normal estimation is solved via direct regression using DCNNs \cite{GB-SNfromRGB_15_Deep3D,GB-SNfromRGB_16_SkipNet,GB-SNfromRGB_19_SR,GB-SNfromRGB_20_TiltedSN,GB-SNfromRGB_21_BAE}. While most methods only estimate the normal, recent work by \cite{GB-SNfromRGB_21_BAE} also estimates the associated uncertainty. They also proposed to apply the training loss on a subset of pixels selected based on the estimated uncertainty.\\
\textbf{Efficient transformer architectures.}
In recent years Transformers \cite{attention_all_need} have been used for a range of different computer vision tasks \cite{vit,video_bert,image_transformer,sceneformer}. However, the all-to-all attention mechanism used in classical transforms suffers from a quadratic scaling problem. \cite{perceiver,perceiver_io,hierachial_perceiver} are one line of work aiming to make transformers more efficient and enabling them to deal with larger inputs. They achieve this by using cross-attention from the inputs to a small set of latent units and subsequently only perform all-to-all attention in this smaller latent space.

\vspace{-0.2cm}
\section{Method}\vspace{-0.2cm}
\begin{figure*}[t]
    \centering
    \includegraphics[width=1.0\linewidth]{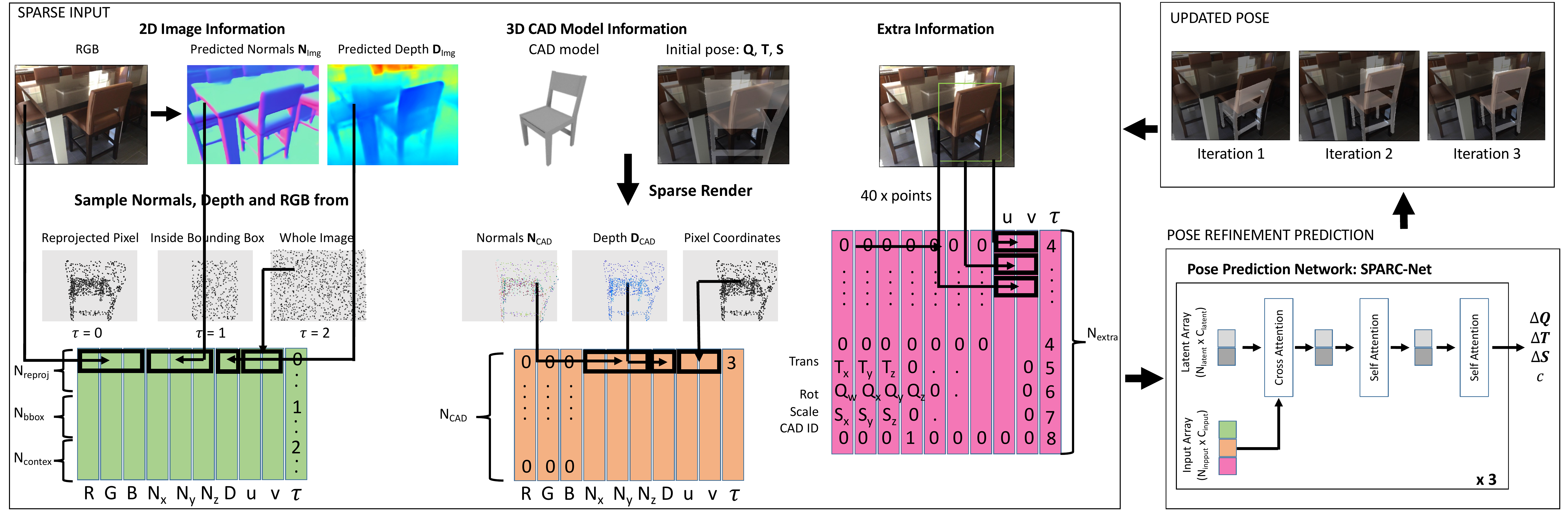}
    \vspace{-0.3cm}
    \caption{\textbf{Method.} 
    (i) Given an RGB image we predict per pixel surface normals $\mathbf{N}_{Img}$ and depth $\mathbf{D}_{Img}$ (2D image information) and use bounding box and CAD model retrievals from \cite{roca} (see Sec. \ref{sec_object_detection}). (ii) We sparsely sample RGB colors, $\mathbf{N}_{Img}$ and $\mathbf{D}_{Img}$ from three different regions (the reprojected CAD model points, the bounding box and the whole image), stack them and add pixel coordinates $(u,v)$ and a token $\tau$ allowing the network to distinguish between different input types. 3D points and surface normals are sampled from the CAD model in an initial canonical pose and reprojected into the image plane (see Sec. \ref{sec_input_fusion}). 2D image information, 3D CAD model information and extra information is combined to form the input to the pose prediction network, SPARC-Net, predicting pose update steps $ \Delta \textbf{T}, \Delta \textbf{R}, \Delta \textbf{S}$ and rotation classification score $c$. Based on the prediction the pose is updated, the CAD model sparsely rendered and the next pose update is predicted (see Sec. \ref{sec_method_pose_update}).}
    \vspace{-0.3cm}
    \label{method_fig}
\end{figure*}
This section explains the key steps of our method. In a first step, we perform object detection as well as surface normal and depth estimation in an image.
In a second step, we sample depth and normal values from an image and combine them with reprojected depth and surface normals sampled uniformly from a CAD model. Finally, in a third step, we iterate the refinements of the 9-DoF initial pose of the CAD model, partially recomputing sparse inputs of step two for every refinement.
\vspace{-0.2cm}
\subsection{Object Detection, Normal and Depth Prediction}
\label{sec_object_detection}
As a first step we perform 2D Object detection, including bounding box and category prediction. The category prediction determines the class of CAD model that is retrieved while the bounding box allows us to initialise the CAD model pose (see Sec. \ref{sec_experimental_setup}) and guide the point sampling (see Sec. \ref{sec_input_fusion}).
We use the same object detections and CAD model retrievals as ROCA \cite{roca} for exact comparability. However, any other method could be used as well. Further we estimate per pixel surface normal $\mathbf{N}_{\text{Img}}$ and depth values $\mathbf{D}_{\text{Img}}$ as these contain crucial geometric information that can be used for precise pose predictions.
% We train a state-of-the-art surface normal estimation network \cite{GB-SNfromRGB_21_BAE} on ground truth surface normals provided by \cite{GB-SNfromRGB_19_FrameNet} and the single-view component of the state-of-the-art depth estimation \cite{Bae_2022_CVPR_depth} on ground truth depth as provided by ScanNet \cite{scannet} (for details see Supp. mat.).
For both surface normal and depth estimation, we use a light-weight convolutional encoder-decoder architecture from \cite{GB-mono-2018-densedepth}. The training losses are the same as the state-of-the-art works \cite{GB-SNfromRGB_21_BAE} for surface normal estimation and \cite{Bae_2022_CVPR_depth} for depth estimation. We use ground truth surface normals provided by \cite{GB-SNfromRGB_19_FrameNet} and ground truth depth as provided by ScanNet \cite{scannet} (for details see Supp. mat.). When training the surface normal and depth network the train and test split used for evaluating SPARC-Net is respected.

% For both tasks, we predict the per-pixel probability distribution for the output and supervise the network by minimizing the negative log-likelihood (NLL) of the ground truth. For surface normal, we parameterize the distribution using the Angular vonMF distribution, proposed in \cite{GB-SNfromRGB_21_BAE}. After training, we discard the uncertainty and only use the predicted mean values.

\vspace{-0.2cm}
\subsection{Input Fusion}
\label{sec_input_fusion}
When attempting to estimate CAD model pose one is presented with two different kinds of information: 2D image information and 3D CAD model information. We fuse the two by sampling 3D points and corresponding surface normals from the CAD model and reprojecting those into the image plane. Comparing the reprojected surface normals and the depth associated with the reprojected point to depth and surface normals estimated from the 2D image allows a network to easily evaluate the current pose and predict a pose update based on this comparison.\\
\textbf{2D image information.}
We stack RGB colors and predicted surface normals $\mathbf{N}_{\text{Img}}$ and depth values $\mathbf{D}_{\text{Img}}$. However, rather than using the entire image as input we sparsely sample pixels from three selected regions. We sample $N_{\text{reproj}}$ pixels onto which CAD model points are reprojected (see paragraph ``3D CAD model information''), $N_{\text{bbox}}$ pixels inside the bounding box and $N_\text{context}$ pixels from the entire image.
Respectively, these provide information on whether the current pose matches the image, what the pose update should be and global context.
In Fig. \ref{iterative_refinement}b and Tab. \ref{table_ablation} we demonstrate that sampling image information sparsely reduces overfitting, leading to a better alignment accuracy, and greatly reducing memory consumption and network complexity (see Sec. \ref{sec_ablation}). For each of the sampled pixels containing color, depth and surface normal values we add the pixel coordinates $(u,v)$ and a token $\tau$ (to allow the network to distinguish between different types of inputs) and stack them to a $(N_{\text{reproj}} + N_{\text{bbox}} + N_\text{context}) \times (3 + 3 + 1 + 2 + 1)$ dimensional tensor as seen in Fig. \ref{method_fig}.\\
\textbf{3D CAD model information.}
We represent the CAD model as a collection of $N_{\text{CAD}}$ 3D points sampled uniformly from the object surface. We simply reproject 3D object points and their surface normals into the image plane which reduces the rendering operation to a perspective projection using a single matrix multiplication. The intuition behind this idea is that for texture-less CAD models almost all information is contained in the 3D shape. Hence there is no need for a full rendering pipeline that takes into account material, lighting, textures or visibility\footnote{While reprojecting all CAD model points into the image plane ignores potentially important occlusion effects we found SPARC-Net to perform well regardless. We have also tried to add information about whether a 3D object point is occluded or not, but did not observe any improvement in performance.}.
We stack reprojected surface normals $\mathbf{N}_{\text{CAD}}$ and the computed depth values $\mathbf{D}_{\text{CAD}}$ with their pixel coordinates and a token $\tau = 3$. Padding this tensor with zeros as RGB values we obtain a $N_\text{CAD} \times 10$ shaped tensor.\\
\textbf{Extra information.}
Additionally, we encode the bounding box information by sampling 10 points from each of the sides of the bounding box and inputting their pixel values  with a different token $\tau = 4$ (padding extra dimensions with 0's). We also directly input the initialised pose consisting of translation $\textbf{T}$, rotation parameterised as quaternion $\textbf{Q}$, scale $\textbf{S}$ and the CAD model ID encoded as a binary vector, each padded with 0's and a unique token. Finally, we concatenate the three different blocks containing image, reprojected CAD model and extra information to form a ($N_{\text{reproj}} + N_{\text{bbox}} + N_\text{context} + N_{\text{CAD}} + N_\text{extra}) \times 10 $  = ($N_\text{input} \times 10)$ dimensional tensor. This tensor is Fourier encoded with 64 frequency bands with a maximum frequency of 1120 (same as original Perceiver \cite{perceiver} applied to point clouds) resulting in a $N_\text{input} \times (10 + 64 * 2 + 1) = N_\text{input} \times C_\text{input}$ tensor which serves as the network input described in the next section.

\vspace{-0.2cm}
\subsection{Pose updates and Iterative Refinement}
\label{sec_method_pose_update}
Combining image information and CAD model information in the image plane allows the network to easily evaluate a given pose and predict a refinement based on this evaluation. We repeat this process $N_\text{iter}$ times allowing the network to refine its own predictions. Specifically, given an initial pose ($\textbf{T},\textbf{Q},\textbf{S}$) and all combined information explained in Sec. \ref{sec_input_fusion} we use a Perceiver network \cite{perceiver} to predict refinement steps ($ \Delta \textbf{T}, \Delta \textbf{Q}, \Delta \textbf{S}$) such that ($ \textbf{T} + \Delta \textbf{T}, \textbf{Q} \cdot \Delta \textbf{Q}, \textbf{S} + \Delta \textbf{S}$) is close to the correct pose $(\textbf{T}_{\text{gt}},\textbf{Q}_\text{gt},\textbf{S}_\text{gt})$ = ($\textbf{T} + {\Delta \textbf{T}}_\text{gt},\textbf{Q} \cdot {\Delta \textbf{Q}}_\text{gt},\textbf{S} + {\Delta \textbf{S}}_\text{gt}$).
% + ( {\Delta \textbf{T}}_\text{gt}, {\Delta \textbf{Q}}_\text{gt}, {\Delta \textbf{S}}_\text{gt}$
Similar to previous works \cite{mask2cad,patch2cad} we find that learning rotations over the full, non-euclidean rotation space is difficult. We therefore follow a coarse-to-fine approach where we simultaneously train the network to predict a binary classification $c$ whether an initial rotation lies within the correct $90^\circ$ rotation bin $c_\text{gt}$ around the vertical.
For classifying rotation we use a standard binary cross-entropy loss $L_{\mathrm{BCE}}$ and for learning the offsets we use the L2 loss
% $L_{\mathrm{L2}} = (\mathbf{x}_\text{target} - \mathbf{x}_\text{pred})^2$
such that our loss function is given by:
\vspace{-0.2cm}
\begin{equation}
\label{loss_equation}
L_{\text {align }}=
w_{\text {c}} L_{\mathrm{BCE}}(c_{\text{gt}},c)
+
w_{\text {t}} L_{\mathrm{L2}}(\Delta \mathbf{T}_\text{gt},\Delta \mathbf{T}) 
+
w_{\text {s}} L_{\mathrm{L2}}(\Delta \mathbf{S}_\text{gt},\Delta \mathbf{S}) 
+
w_{\text {q}} L_{\mathrm{L2}}(\Delta \mathbf{Q}_\text{gt},\Delta \mathbf{Q}).
\end{equation}\vspace{-0.4cm}

% At test time we binary classify four rotation initialisation that are $90^\circ$ rotated around the vertical. For the initialisation with the highest estimated probability $c$ we subsequently predict the offsets and iteratively rerender and predict again.
At test time we apply SPARC-Net to four rotation initialisation that are $90^\circ$ rotated around the vertical. For each of those we predict the probability $c$ indicating whether the correct rotation lies within $\pm 45^\circ$ of the tested initialisation. For the initialisation with the highest estimated probability $c$ we subsequently predict the 9 DoF pose updates and iteratively rerender and refine the pose.

\vspace{-0.2cm}
\section{Experimental Setup}
\vspace{-0.2cm}
\label{sec_experimental_setup}
This section briefly describes the dataset used for training and testing our method, as well as the evaluation metrics used and the hyperparameters chosen.\\
\textbf{ScanNet dataset.}
Similar to \cite{mask2cad,patch2cad,total_3D_understanding,roca} we train and test our approach on the ScanNet25k image data \cite{scannet} for which \cite{scan2cad} provide CAD model annotation for a wide range of objects. This dataset contains 20k training images representing 1200 train scenes and 5k validation images from 300 different evaluation scenes. We train and test our method on the 9 categories with the most CAD model annotations covering more than 2500 distinct shapes.\\
\textbf{Evaluation metric.}
We follow the original evaluation protocol introduced by Scan2CAD \cite{scan2cad} which evaluates CAD model alignments per-scene. In the same way as ROCA \cite{roca} we transform predicted CAD model poses into ScanNet \cite{scannet} world coordinates and apply 3D non-maximum suppression to eliminate multiple detections of the same object from different images. A CAD model prediction is considered correct if the object class prediction is correct, the translation error is less than 20 cm, the rotation error is less than $20^\circ$ and the scale ratio is less than $20\%$.
% CAD model predictions can be made up to the total number of ground truth CAD models per scene.
We found that there was a bug in the original evaluation code introduced by \cite{scan2cad} which was subsequently used for evaluating \cite{roca,mask2cad}. When computing the scale error the formula $\text{s}_{\text{error}} = | \sum_{i=x,y,z} ( S_i / S^{\text{gt}}_i  - 1 )|$ was used instead of $\text{s}_{\text{error}} = \sum_{i=x,y,z}  | (S_i / S^{\text{gt}}_i)  - 1 |$ which allowed scale errors in different directions to cancel each other out. We corrected this mistake and reevaluated \cite{mask2cad,roca} (see Supp. mat.).\\
\textbf{Input details.}
For the main experiment we choose
$N_\text{CAD}=1000$ and use the reprojected points as samples from the 2D image such that $N_\text{reproj}=1000$. Further we use 
$N_\text{bbox}=1000$ and $N_\text{context}=5000$ as also shown in Fig. \ref{method_fig}.\\
\textbf{Perceiver architecture.}
For the Perceiver \cite{perceiver} we set $N_\text{latent} = 128$ and $C_\text{latent} = 256$ and as depicted in Fig. \ref{method_fig} repeat three blocks of one cross attention layer followed by two self-attention layers with weight sharing between each of the layers in the three blocks.\\
\textbf{Train details}.
We train the SPARC-Net for 300 epochs using the LAMB \cite{lamb} optimiser (as used by the original Perceiver \cite{perceiver}) with learning rate 0.001. For the loss function in Eq.~\ref{loss_equation} we set $w_c = 0.5$, $w_t = 0.5$, $w_s = 0.5 $ and $w_q = 1$. We use a batchsize of 80. For more information on sampling and initialising poses at train and test time see the Supp. mat.\\
\textbf{Implementation Details}.
All code is implemented in PyTorch \cite{pytorch}. SPARC-Net was trained on a single TitanXp for 36 hours.
\vspace{-0.2cm}
\section{Experimental Results}
In the first part of this section we compare ourselves to the state-of-the-art approaches \cite{roca,mask2cad,total_3D_understanding}. In the second part, we investigate and ablate different key aspects of our system, including the 2D pixel sampling, the sparse 3D object representation, render-and-compare, the number of refinements used and the relevance of different input information.
\begin{figure*}[t]
    \centering
    \includegraphics[width=1.0\linewidth]{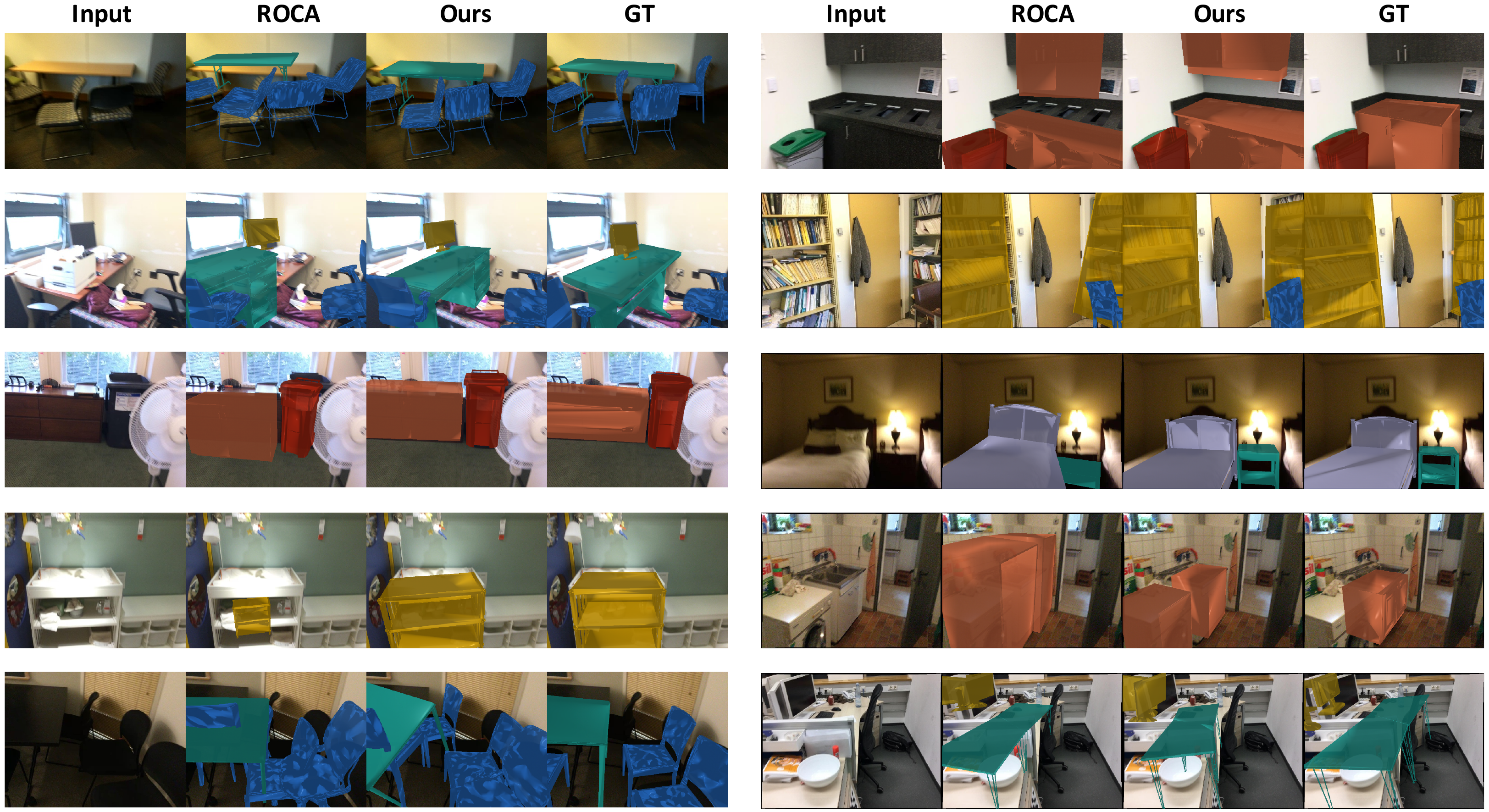}\vspace{-0.3cm}
    \caption{\textbf{Qualitative results on ScanNet \cite{scannet}.} Our sparse, render-and-compare approach allows for more precise CAD model alignment compared to ROCA \cite{roca}. Typical failure cases of ROCA include systematic failure when predicting normalised object coordinates leading to wrong object translations (row 3) and wrong scale predictions (row 4). Row 5 shows failure cases for both methods in complex scenes (left) and when object boundaries are not clearly identifiable (right).}\vspace{-0.3cm}
    \label{qualitative_results}
\end{figure*}

\vspace{-0.3cm}
\subsection{Main Results}
 Tab. \ref{table_main_results} shows that our approach outperforms all competing approaches in all object classes (except for ROCA on the class ``display''). Further we significantly improve both the class average alignment accuracy from 18.4\% to 24.9\% as well as the instance alignment accuracy from 25.0\% to 31.8\%. Visually comparing our predictions to ROCA \cite{roca} in Fig. \ref{qualitative_results} we note that the improvements are mainly due to better translation and scale predictions. ROCA \cite{roca} relies on predicting normalised object coordinates for estimating translation. However, due to difficulties in aligning and registering different CAD models to a canonical frame, predicting NOCs often fails with a systematic offset leading to an offset in the final alignment (see Supp. mat.). In contrast, our approach is independent of the existence of a canonical object frame as it directly compares a reprojected object to an image. Further, ROCA decouples the scale predictions from predicting rotation and translation which can produce very wrong alignments for bad scale predictions (see Fig. \ref{qualitative_results}).
 We also evaluate our system when using ROCA's rotation predictions as an initialisation as opposed to our own classified rotation bins. Here we observe a further improvement of our alignment accuracy. The reason why ROCA's rotation predictions are good is that they are computed geometrically from correspondences and are largely unaffected by systematic offsets in NOCs predictions and wrong scale predictions.

\renewcommand{\arraystretch}{1.1}

% \begin{table}[t]
%     \centering
%     \resizebox{\textwidth}{!}{
%     \begin{tabular}{S|SSSSSSSSS|SS}
%     % {||p{3cm} | p{0.8cm} p{0.8cm} p{0.8cm} p{0.8cm} p{0.8cm} p{0.8cm} p{0.8cm} p{0.8cm} p{0.8cm} p{0.8cm}||} 
    
%      Method & bathtub & bed & bin & bkshlf & cabinet & chair & display & sofa & table & \textbf{class} & \textbf{instance}\\ [0.5ex] 
%      \hline
%     % \toprule
%      Number of Instances \# & 120 & 70 & 232 & 212 & 260 & 1093 & 191 & 113 & 553 & 9 & 2844 \\
%      \hline
%     % \midrule
%      Total3D-ODN \cite{total_3D_understanding}& 10.0 & 2.9 & 16.8 & 2.8 & 4.2 & 14.4 & 13.1 & 5.3 & 6.7 & 8.5 & 10.4 \\
%      Mask2CAD-b5 \cite{mask2cad} & 7.5 & 2.9 & 24.6 & 1.4 & 5.0 & 29.9 & 13.1 & 5.3 & 5.6 & 10.6 & 16.7 \\ 
%      ROCA \cite{roca} & 20.8 & 8.6 & 26.3 & 9.0 & 13.1 & 39.9 & 24.6 & 10.6 & 12.7 & 18.4 & 25.0\\
%      \hline
%      Ours & 13.3 & 31.4 & 30.6 & 11.8 & 18.1 & 46.7 & 22.5 & 18.6 & 17.5 & 23.4 & 30.0
%       \\
%      Ours + ROCA rot init & 25.0 & 25.7 & 38.8 & 14.2 & 20.4 & 53.0 & 25.1 & 23.9 & 18.6 & 27.2 & 34.4\\ [1ex] 
%     \end{tabular}}
%     \caption{\textbf{Alignment Accuracy on ScanNet} \cite{scannet,scan2cad} in comparison to the state-of-the-art.}
%     \label{table_main_results}
% \end{table}

\vspace{-0.2cm}
\subsection{Ablation}
\label{sec_ablation}
Note that in addition to average class alignment accuracy and instance alignment accuracy we report $\mathbf{T},\mathbf{R},\mathbf{S}$ and $c$ accuracy in Tab. \ref{table_ablation} where the same evaluation protocol is used as described in Sec. \ref{sec_experimental_setup}, but rather than requiring $\mathbf{T}$, $\mathbf{R}$ and $\mathbf{S}$ to be correct at the same time only the quantity of interest is required to be correct ($c$ is considered correct if the binary rotation classification is correct). If not otherwise specified in the following ``accuracy'' or ``performance'' refers to instance alignment accuracy.

\begin{figure*}[t]
    \centering
    \includegraphics[width=1.0\linewidth]{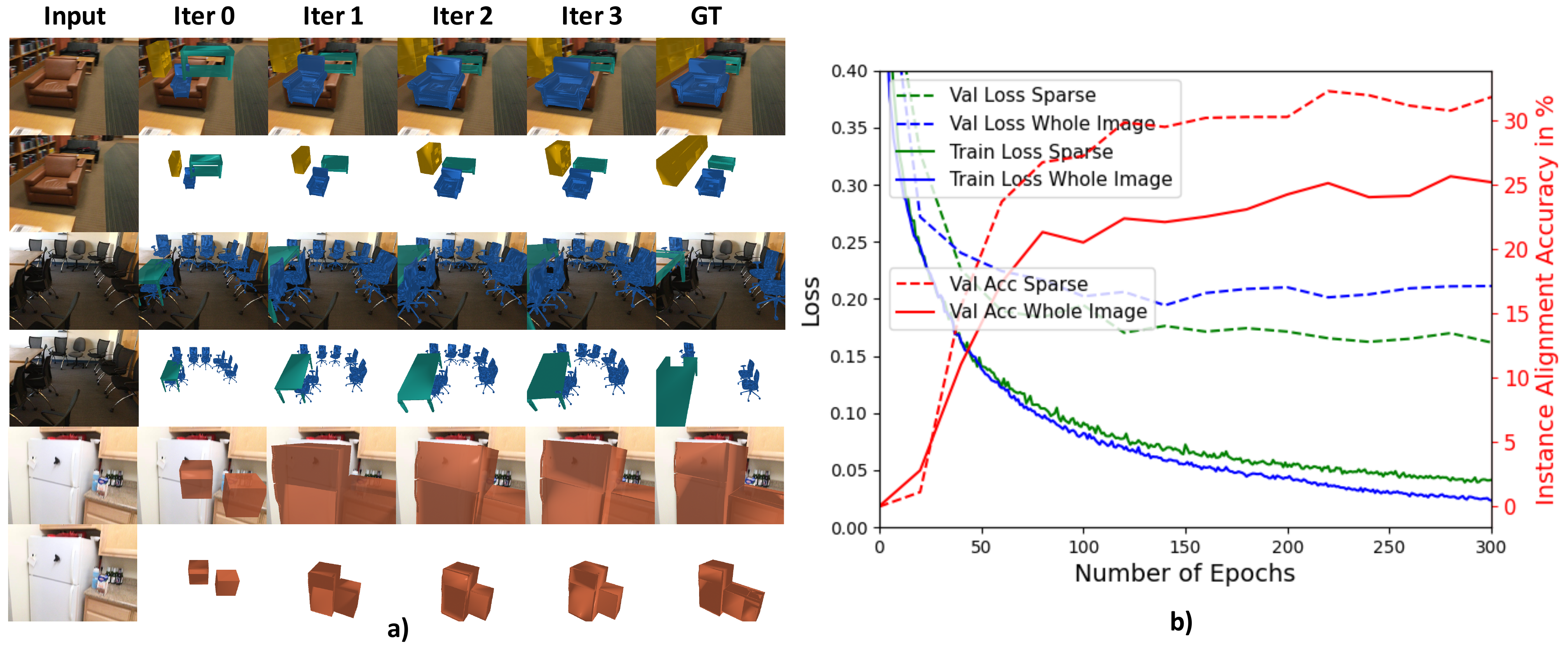}\vspace{-0.3cm}
    \caption{ a) \textbf{Visualisation of refinements} in iterative render-and-compare. b) \textbf{Train and validation loss as well as validation accuracy} on the ScanNet dataset \cite{scannet,scan2cad}. When training on the whole image SPARC-Net overfits the training data more easily resulting in higher validation loss and worse instance alignment accuracy.}\vspace{-0.1cm}
    \label{iterative_refinement}
\end{figure*}

\begin{table}[t]
    \centering
    \resizebox{\textwidth}{!}{
    \begin{tabular}{c|ccccccccc|cc}
    % {||p{3cm} | p{0.8cm} p{0.8cm} p{0.8cm} p{0.8cm} p{0.8cm} p{0.8cm} p{0.8cm} p{0.8cm} p{0.8cm} p{0.8cm}||} 
    
     Method & bathtub & bed & bin & bkshlf & cabinet & chair & display & sofa & table & \textbf{class} & \textbf{instance}\\ [0.5ex] 
     \hline
    % \toprule
     Number of Instances \# & 120 & 70 & 232 & 212 & 260 & 1093 & 191 & 113 & 553 & 9 & 2844 \\
     \hline
    % \midrule
     Total3D-ODN \cite{total_3D_understanding}& 10.0 & 2.9 & 16.8 & 2.8 & 4.2 & 14.4 & 13.1 & 5.3 & 6.7 & 8.5 & 10.4 \\
     Mask2CAD-b5 \cite{mask2cad} & 7.5 & 2.9 & 24.6 & 1.4 & 5.0 & 29.9 & 13.1 & 5.3 & 5.6 & 10.6 & 16.7 \\ 
     ROCA \cite{roca} & 20.8 & 8.6 & 26.3 & 9.0 & 13.1 & 39.9 & \textbf{24.6} & 10.6 & 12.7 & 18.4 & 25.0\\
     \hline
    %  Ours & 24.2  & 25.7 & 25.0 & 12.3 & 18.1 & 50.1 & 13.6 & 21.2 & 15.2 & 22.8 & 30.2
    %   \\
    SPARC-Net (ours) & \textbf{25.8} & 25.7 & 24.6 & \textbf{14.2} & \textbf{20.8} & 51.5 & 17.8 & \textbf{28.3} & 15.4 & 24.9 & 31.8 \\
    SPARC-Net + ROCA rot init & 25.0 & \textbf{30.0} & \textbf{36.2} & \textbf{14.2} & 19.2 & \textbf{52.3} & 20.4 & \textbf{28.3} & \textbf{20.1} & \textbf{27.3} & \textbf{34.1}\\
    %  Ours + ROCA rot init & 25.0 & 25.7 & \textbf{38.8} & \textbf{14.2} & 20.4 & \textbf{53.0} & \textbf{25.1} & 23.9 & \textbf{18.6} & \textbf{27.2} & \textbf{34.4}\\
    %  [1ex] 
    \end{tabular}}
    \caption{\textbf{Alignment Accuracy on ScanNet} \cite{scannet,scan2cad} in comparison to the state-of-the-art.}\vspace{-0.5cm}
    \label{table_main_results}
\end{table}
\noindent
\textbf{Image sampling.}
We compare sampling the entire image to sparse sampling and observe worse alignment accuracy when using the entire image ($25.0\%$ vs. $31.8\%)$ due to faster overfitting at train time (see Fig. \ref{iterative_refinement}b). We further simplify our sampling scheme and instead of sampling the image globally, the bounding box and the reprojected query, we use samples from just the bounding box, $N_{\text{bbox}} = 1000$. Surprisingly, we observe that we can recover almost the same alignment accuracy $30.1\%$ compared to $31.8\%$ before.
By traininig and testing with less than 1\% of the original number of pixel (image resolution 480 x 360) we reduce the memory by a factor of 100.
% As all images are processed at resolution 480x360 this demonstrates that we can effectively train and test with less than 1\% of the original number of pixels.
% We therefore not only \textcolor{red}{reduce the memory at train and test time by a factor of 100}  but as
The complexity of the Perceiver \cite{perceiver} is given by 
$\mathcal{O}(N_\text{input} N_\text{latent}) + \mathcal{O} (L N_\text{latent}^{2}) \approx \mathcal{O} (N_\text{input} N_\text{latent})$  where $N_\text{input} >> N_\text{latent}$, $N_\text{input} >> L$. When using the whole image as input $N_\text{input}$ is dominated by the number of pixels $N_\text{input} \approx 480 \times 360$. Therefore as $N_\text{latent} = 128$ and $L = 2$ (two self-attention layers following the cross-attention layer) sparse inputs also reduce the compute complexity by a factor of 100, greatly speeding up training and testing.\\
\textbf{Sparse CAD model representation.}
When reducing the number of points $N_{\text{CAD}}$ we observe that remarkably with just 100 points we obtain similar performance as with 1000 points (30.0\% vs. 31.8\%), showing that 3D shapes can indeed be represented very sparsely. However, as we further decrease the number of points we observe a rapid decline in performance (20.9\% for $N_\text{CAD} = 50$ and $4.2\%$ for $N_\text{CAD} = 50$. Interestingly, this decrease is almost entirely due to worse rotation predictions (see $\mathbf{R}$ accuracy $24.8\%$ compared to $67.9\%$). Intuitively, this makes sense as an extremely sparse point cloud has very little structure from which to predict rotation while translation and scale can still be predicted. Initialising those extremely sparse models with ROCA \cite{roca} rotation predictions we can still recover reasonable alignment accuracy of $24.0\%$ and $29.9\%$ for $N_\text{CAD} = 20$ and $N_\text{CAD} = 50$.\\
\textbf{Render-and-Compare.}
We test the assumption that Render-and-Compare is beneficial for precise pose predictions by considering two networks which are trained either on no 3D shape information but just the CAD model ID or for which the CAD model is always presented in the same canonical pose. The first is motivated by works (e.g. \cite{mask2cad,patch2cad}) that contain no explicit 3D shape information and the second by \cite{kanazawaHMR18} that perform iterative pose updates without explicit rerendering. Similar to before both of these networks fail to learn accurate rotation predictions (both for the classification and the refinement) therefore we only show results where we initialise poses with ROCA \cite{roca} rotation predictions. Here we note that using just the CAD ID results in a significant decrease in performance (21.0\%), whereas using 3D information but presenting the CAD model in the same canonical pose achieves $25.9\%$ instance alignment accuracy.\\
\textbf{Number of refinements.}
Training and evaluating our network for different number of iterations we observe a significant improvement when increasing the number of refinements from 1 ($26.4\%$) to 3 ($31.8\%$) and no more improvement for 5 refinements ($31.3\%$).

\begin{table}[t]
    \centering
    \resizebox{\textwidth}{!}{
    \begin{tabular}{c|ccccc|cccccc|cc|ccc|cccc}
    
    \textbf{Accuracy} & \multicolumn{5}{c}{\textbf{2D Image Sampling}} & \multicolumn{6}{c}{\textbf{3D CAD model points}} & \multicolumn{2}{c}{\textbf{R. and C.}} &\multicolumn{3}{c}{\textbf{N Refinement}} & \multicolumn{4}{c}{\textbf{Input Information}} \\
    \hline
& \makecell{$N_\text{reproj}=100$\\$N_\text{bbox}=100$\\$N_\text{context}=0$\\$N_\text{CAD}=100$} & \makecell{$N_\text{reproj}=0$\\$N_\text{bbox}=200$\\$N_\text{context}=0$\\$N_\text{CAD}=100$}& \makecell{$N_\text{reproj}=0$\\$N_\text{bbox}=1000$\\$N_\text{context}=0$\\$N_\text{CAD}=100$} & \makecell{$N_\text{reproj}=0$\\$N_\text{bbox}=0$\\$N_\text{context}=5000$\\$N_\text{CAD}=1000$} & \makecell{Whole img \\$N_\text{CAD}=1000$} & \makecell{20} & \makecell{20*} & \makecell{50} & \makecell{50*} & \makecell{100} & 1000 & \makecell{Just\\CAD\\ID*} & \makecell{Canon-\\ical\\pose*} & 1 & 3 & 5 & \makecell{No\\D} & \makecell{No\\N} & \makecell{No\\RGB} & \makecell{No RTS\\ or CAD\\ info} \\
\hline
\textbf{Class} & 12.9 & 12.7 & 23.9 & 14.8 & 20.3 & 3.5 & 17.7 & 15.9 & 24.9 & 23.1 & 24.9 & 15.0 & 20.2 & 16.6 & 24.9 & 25.2 & 15.0 & 16.6 & 24.7 & 22.2 \\
\rowcol \textbf{Instances} & 18.5 & 19.7 & \textbf{30.1} & 20.1 & \textbf{25.0} & \textbf{4.2} & \textbf{24.0} & \textbf{20.9} & \textbf{29.9} & \textbf{30.0} & \textbf{31.8} & \textbf{21.0} & \textbf{25.9} & \textbf{26.4} & \textbf{31.8} & \textbf{31.3} & \textbf{18.8} & \textbf{20.7} & \textbf{30.5} & \textbf{28.9} \\
\textbf{T} & 40.8 & 42.1 & 42.4 & 37.1 & 38.7 & 38.0 & 39.9 & 38.2 & 42.5 & 42.2 & \textbf{45.4} & 36.4 & 41.4 & 42.3 & \textbf{45.4} & 43.7 & \textbf{30.7} & 40.2 & 42.6 & \textbf{42.2} \\
\textbf{R} & 49.1 & 48.7 & 70.8 & 61.4 & 68.2 & \textbf{24.8} & 63.7 & 58.4 & 70.9 & 71.1 & \textbf{67.9} & 63.8 & 63.7 & 65.5 & \textbf{67.9} & 71.5 & 68.2 & \textbf{56.8} & 72.3 & 66.3 \\
\textbf{S} & 61.0 & 62.4 & 69.0 & 61.3 & 66.8 & 63.7 & 63.3 & 66.4 & 67.7 & 67.8 & 68.4 & 62.9 & 68.9 & 62.8 & 68.4 & 66.4 & 64.5 & 63.7 & 68.5 & 67.2 \\
\textbf{R class} & 70.7 & 71.3 & 75.2 & 69.7 & 72.9 & 53.4 & 76.4 & 68.9 & 77.4 & 75.6 & 75.9 & 77.0 & 76.3 & 74.8 & 75.9 & 75.6 & 73.0 & 74.5 & 76.1 & 74.4
\end{tabular}}

%  \textbf{Model} VGG16 16.9  \textbf{25.8}  & 43.7 58.4 63.9  & 73.7
\caption{\textbf{Ablation.} We perform a number of different ablations as explained in detail in Sec. \ref{sec_ablation}. Bold numbers are the ones referred to in Sec. \ref{sec_ablation}. Note that a ``*'' indicates that ROCA \cite{roca} rotation predictions are used for initialising rotations. Our main experiment is listed twice under ``3D CAD model points $N_\text{CAD} = 1000$'' and ``N Refinements, $N = 3$'' for convenience. ``R. and C.'' stands for Render-and-Compare.}\vspace{-0.4cm}
\label{table_ablation}
\end{table}
\noindent
\textbf{Input modalities.}
Furthermore, we investigate the importance of the different information that is provided as input to the network. Here we find that both depth and normal predictions are crucial for SPARC-Net. Not providing depth predictions from the 2D image reduces the accuracy to $18.8\%$ while not providing normal predictions reduces it to $20.7\%$. We find that for depth this drop in performance is mainly due to worse $\mathbf{T}$ predictions ($30.7\%$ compared to $45.4\%$) whereas for the normals it is due to worse rotation predictions ($56.8\%$ compared to $67.9\%$) which is exactly as one would expect. In contrast, not providing RGB color only reduces the accuracy marginally to $30.5\%$. This makes sense as given depth and surface normal estimates color only adds very little information for shape alignments. When no extra information is provided the accuracy is reduced to $28.9\%$. The individual accuracies suggest that the network uses extra information to learn the distribution of object positions as we observe the largest decrease for $\mathbf{T}$ accuracy to $42.2\%$.\\
% \textbf{Model.}
% Finally, we replace our Perceiver model with a conventional VGG16 \cite{vgg} which is trained on entire images and observe a drop in accuracy to $25.8\%$. \textcolor{red}{Actually the perceiver trained on whole images also just has an accuracy of $25.0\%$.}
% However, for more than one refinement steps the inputs differ as the Render-and-Compare approach sees the CAD model reprojected in the updated pose whereas the No-Update network only receives information about the updated pose directly from the values of R, S and T in the input. As can be observed these by themselves do not provide enough information for the network to benefit from an iterative refinement procedure and the No-Update method does not improve with additional refinements whereas the Render-and-Compare model does. 

\vspace{-0.6cm}
\section{Conclusion}\vspace{-0.1cm}
We have proposed SPARC, a sparse render-and-compare approach for CAD model alignment from a single RGB image. We show how to effectively combine 3D CAD model information and 2D image information with early-fusion which helps our pose prediction network to estimate accurate pose updates. In this way we improve the state-of-the-art on ScanNet~\cite{scannet} from $25.0\%$ to $31.8\%$ instance alignment accuracy.
% Further we highlight fundamental issues when attempting to register differently shaped CAD models with each other which can lead to systematic offsets when attempting to estimate their normalised object coordinates.
We demonstrate that using sparse 2D image information (less than 1\% of available pixels) reduces overfitting leading to better pose predictions and greatly reducing memory and network complexity. Furthermore, sparse input processing may bring even more benefits when dealing with sensory input that is naturally sparse such as LIDAR or event cameras.
% While CAD models retrieved from a database can provide a clean and compact scene representation, they can not perfectly match all object in a scene. 
In the future, we plan to expand this work to updating not only the estimated pose, but also the CAD model shape in order to alleviate the requirement of CAD models that are very similar to the objects in the image. 
% match more closely objects that are not present in the databa

\bibliography{bmvc}

\newpage
\section*{Supplementary Material}
\appendix

We provide additional information for various aspects of our main work. 
In Sec. \ref{sec_correction} we present alignment accuracies evaluated with the orignal (wrong) evaluation code and the corrected one. In Sec. \ref{sec_additional_train_and_test} and Sec. \ref{sec_train_surface_and_depth} we give additional information for training our pose prediction network, SPARC-Net, and the networks used for depth and surface normal estimation. In Sec. \ref{sec_alinging_cad_models} we highlight issues when trying to align differently shaped CAD models with each other and the resulting systematic offsets that appear in the predictions of the normalised object coordinates. We quantitatively support this section by ablating our system with ROCA \cite{roca} predictions in Sec. \ref{sec_ablating_roca_predictions}. Finally, we discuss limitations and possibility for future works in Sec. \ref{sec_limitations} and explain the provided video showing qualitative results on ScanNet \cite{scannet} in Sec. \ref{sec_vis_video}.
\section{Correction to Evaluation Script}
\label{sec_correction}
Scan2CAD \cite{scan2cad} proposed to consider a CAD alignment correct if the object class prediction is correct, the translation error is less than 20 cm, the rotation error is less than 20 degrees and the scale ratio is less than 20 \%. We found that there was a bug in the original evaluation code which was subsequently used to evaluate ROCA \cite{roca} and Mask2CAD \cite{mask2cad}. When computing the scale ratio the formula $\text{s}_{\text{error}} = | \sum_{i=x,y,z} ( s^{\text{pred}}_i / s^{\text{gt}}_i  - 1 )|$ was used instead of $\text{s}_{\text{error}} = \sum_{i=x,y,z}  | (s^{\text{pred}}_i / s^{\text{gt}}_i)  - 1 |$ which allowed scale errors in different directions to cancel each other out. We correct for this mistake and reevaluate \cite{mask2cad,roca}. The accuracy computed with the evaluation code containing the mistake are presented in Table \ref{table_wrong_alignment}  and their corrected counterparts are presented in Table \ref{table_correct_alignment}.

\section{Additional Train and Test Information}
\label{sec_additional_train_and_test}

\textbf{Train data generation.}
We generate per-image CAD model pose annotations for all images in ScanNet25k \cite{scannet} (which is the original ScanNet\cite{scannet} dataset sampled for every 100th frame) by transforming CAD model pose annotations from \cite{scan2cad} from ScanNet \cite{scannet} world coordinates into camera coordinates. ScanNet25k \cite{scannet} contains ca. 20K train images from 1200 different scenes. Note that for a given image we only train our CAD model to align CAD models whose center is reprojected into the image (such as to avoid training on objects that are barely visible). Further, we filter objects which do not have at least 50\% of their reprojected depth values within 30 cm of the ground truth depth values. This avoids training on objects that are hidden behind walls or in other way strongly occluded.\\
\textbf{Pose sampling at train time.}
When initialising poses for training we uniformly sample translations $\textbf{T}$ to be between 1 and 5 m along the $Z$-axis and sample $X$ and $Y$ by uniformly sampling pixel coordinates in the predicted bounding box and mulitplying the corresponding pixel bearing by the sampled $Z$ to obtain $X$ and $Y$ values.  Scale values $\textbf{S}$ are sampled uniformly in the range of all observed CAD model scales for the detected category. 
25\% of train examples are uniformly sampled in random rotation (any azimuthal angle, $20^\circ$ tilt range and $40^\circ$ elevation range around $\textbf{R}^\text{gt}$) to learn to classify rotations, whereas 75\% of train examples are sampled in the correct rotation bin ($90^\circ$ azimuthal angle range, $20^\circ$ tilt range and $40^\circ$ elevation range around $\textbf{R}^\text{gt}$) to learn to regress pose offsets.\\
At test time the pose prediction network is used to iteratively refine its own predictions. During training it is therefore crucial to not just generate random CAD model poses, but also poses based on the networks predictions. This ensures that the poses sampled during training are as similar to the ones the network sees at test time as possible. For every image and corresponding CAD model annotation we therefore use a randomly initialised pose as well as the two subsequent refinements predicted by the network as training data. Specifically, for a batch of $N$ train examples, each containing image information and CAD model information sampled in a random initial pose, we predict the pose updates and apply losses. Based on the predicted pose updates, we update the CAD model poses for every train example and recompute the inputs. The recomputed inputs are fed through the network again, pose updates are predicted and losses applied. This process is repeated once more, after which new training images with new CAD models in random initial poses are sampled as the next training examples. \\
% When training the alignment network it is crucial to sample poses that are similar to the ones seen during the iterative refinement process at test time. We ensure this by using the predicted pose update to reproject points sampled from all CAD models in the current batch and using those as the next train examples. After repeating this two times we sample different images again and initialise the objects with random poses.\\
\textbf{Test details.}
At test time $\textbf{R}$ is initialised with four rotations at 0, 90, 180 and 270 degrees around the vertical axis, 0 degrees tilt and 20 elevation angle (such that the camera is looking slightly down at an object that is standing straight upright). $\textbf{T}$ is initialised to have $z=3$ m and x and y such that the reprojected $\textbf{T}$ lies at the bounding box center. The scale $\textbf{S}$ is initialised with the median value of all CAD models for the given category. For the rotation with the highest classification score $c$ we predict ($ \Delta \textbf{T}, \Delta \textbf{R}, \Delta \textbf{S}$) and iteratively refine the pose 3 times. Note that our pose prediction network is very robust to poor initialisation for scale and translation (see the video explained in Sec. \ref{sec_vis_video}) but can not reorient CAD models if their are initialised within the wrong $90^\circ$ rotation bin.

\section{Details for Training Surface Normal and Depth Networks}
\label{sec_train_surface_and_depth}
For both surface normal $\mathbf{N}_{\text{Img}}$ and depth estimation $\mathbf{D}_{\text{Img}}$, we use a light-weight convolutional encoder-decoder architecture from \cite{GB-mono-2018-densedepth}. For both tasks, we predict the per-pixel probability distribution for the output and supervise the network by minimizing the negative log-likelihood (NLL) of the ground truth. For surface normal, we parameterize the distribution using the Angular vonMF distribution, proposed in \cite{GB-SNfromRGB_21_BAE}, while we parameterise the depth distribution with a gaussian distribution. After training, we discard the uncertainty and only use the predicted mean values. We use ground truth surface normals provided by \cite{GB-SNfromRGB_19_FrameNet} and ground truth depth as provided by ScanNet \cite{scannet}, respecting the train/test split. For depth we train on all two million available train images, while for surface normals we train on all images for which \cite{GB-SNfromRGB_19_FrameNet} provide annotations and that are within the set of train images which results in ca. 200K train images. We train both networks for ten epochs using the AdamW optimiser \cite{adamw} and schedule the learning rate using 1cycle policy \cite{cyclic_lr} with $lr_\text{max} = 3.5 \times 10^{-4}$ (same as \cite{Bae_2022_CVPR_depth}). We use a batch size of four for both surface normal and depth training.

\renewcommand{\arraystretch}{1.1}
\begin{table}[t]
    \centering
    \resizebox{\textwidth}{!}{
    \begin{tabular}{c|ccccccccc|cc}
    % {||p{3cm} | p{0.8cm} p{0.8cm} p{0.8cm} p{0.8cm} p{0.8cm} p{0.8cm} p{0.8cm} p{0.8cm} p{0.8cm} p{0.8cm}||} 
    
     Method & bathtub & bed & bin & bkshlf & cabinet & chair & display & sofa & table & \textbf{class} & \textbf{instance}\\ [0.5ex] 
     \hline
     Mask2CAD-b5 \cite{mask2cad} & 8.3 & 2.9 & 25.9 & 3.8 & 5.4 & 30.9 & 17.3 & 5.3 & 7.1 & 11.9 & 17.9 \\ 
     ROCA \cite{roca} & 22.5 & 10.0 & 29.3 & 14.2 & 15.8 & 41.0 & 30.4 & 15.9 & 14.6 & 21.5 & 27.4\\
     
     Ours & 26.7 & 25.7 & 26.7 & 17.5 & 23.8 & 52.6 & 22.5 & 32.7 & 17.7 & 27.3 & 33.9\\
     Ours + ROCA rot init & 27.5 & 30.0 & 41.4 & 17.5 & 23.5 & 53.7  & 26.2 & 32.7 & 22.6 & 30.6 & 36.8\\
    %  Ours & 27.5  & 27.1 & 28.9  & 16.0  & 21.2  & 50.9  & 20.4 & 23.9 & 16.8  & 25.9  & 32.5
    %   \\
    %  Ours + ROCA rot init & 27.5  & 28.6 & 40.9 & 16.5 & 23.1 & 52.7 & 32.5 & 28.3 & 20.3 & 30.0  & 36.0 \\ [1ex] 
    \end{tabular}}
    \caption{\textbf{Alignment accuracy with original (incorrect) treatment of scale predictions} on ScanNet \cite{scannet}.}
    \label{table_wrong_alignment}
\end{table}

\begin{table}[t]
    \centering
    \resizebox{\textwidth}{!}{
    \begin{tabular}{c|ccccccccc|cc}
    % {||p{3cm} | p{0.8cm} p{0.8cm} p{0.8cm} p{0.8cm} p{0.8cm} p{0.8cm} p{0.8cm} p{0.8cm} p{0.8cm} p{0.8cm}||} 
    
     Method & bathtub & bed & bin & bkshlf & cabinet & chair & display & sofa & table & \textbf{class} & \textbf{instance}\\ [0.5ex] 
     \hline
     Mask2CAD-b5 \cite{mask2cad} & 7.5 & 2.9 & 24.6 & 1.4 & 5.0 & 29.9 & 13.1 & 5.3 & 5.6 & 10.6 & 16.7 \\ 
     ROCA \cite{roca} & 20.8 & 8.6 & 26.3 & 9.0 & 13.1 & 39.9 & 24.6 & 10.6 & 12.7 & 18.4 & 25.0\\
     Ours & 25.8 & 25.7 & 24.6 & 14.2 & 20.8 & 51.5 & 17.8 & 28.3 & 15.4 & 24.9 & 31.8 \\
    Ours + ROCA rot init & 25.0 & 30.0 & 36.2 & 14.2 & 19.2 & 52.3 & 20.4 & 28.3 & 20.1 & 27.3 & 
    34.1\\
    %  Ours & 13.3 & 31.4 & 30.6 & 11.8 & 18.1 & 46.7 & 22.5 & 18.6 & 17.5 & 23.4 & 30.0
    %   \\
    %  Ours + ROCA rot init & 25.0 & 25.7 & 38.8 & 14.2 & 20.4 & 53.0 & 25.1 & 23.9 & 18.6 & 27.2 & 34.4\\ [1ex] 
    \end{tabular}}
    \caption{\textbf{Alignment accuracy with correct treatment of scale predictions} on ScanNet \cite{scannet}.}
    \label{table_correct_alignment}
\end{table}

\section{Aligning CAD Models for Normalised Object Coordinates}
\label{sec_alinging_cad_models}
\begin{figure*}[!h]
    \centering
    \includegraphics[width=1.0\linewidth]{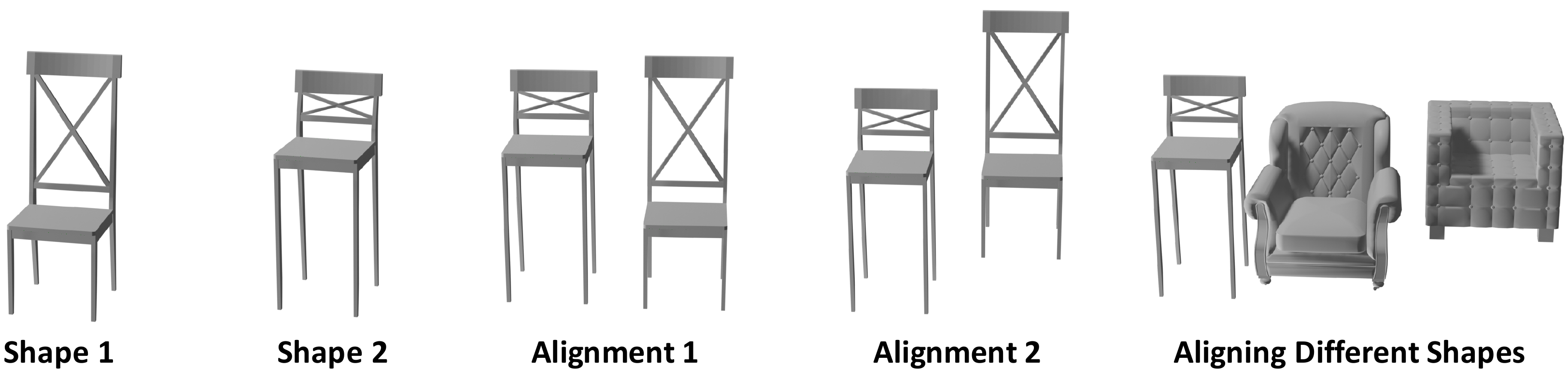}
    \vspace{-0.5cm}
    \caption{\textbf{Issues with shape alignment.} When trying attempting to align Shape 1 and Shape 2 with each other different alignments are possible. Alignment 1 aligns the top and the bottom of the chairs. This ensures that both shapes fit into the same normalised 3D bounding box and is the alignment used for NOCs. Alignment 2 aligns the chairs by their seating area. While now both shapes do not lie in the same normalised 3D bounding box, their seating areas align which is useful when trying to predict their coordinates. The point is that for different shapes different alignments are possible depending on which object parts one wishes to align. The more different the shapes are the harder it is to find some global alignment that aligns all different object parts with each other. This means that NOCs learned for one shape do not generalise well to NOCs learned for other shapes.}
    \label{issue_alignment}
    \vspace{1cm}
    \centering
    \includegraphics[width=1.0\linewidth]{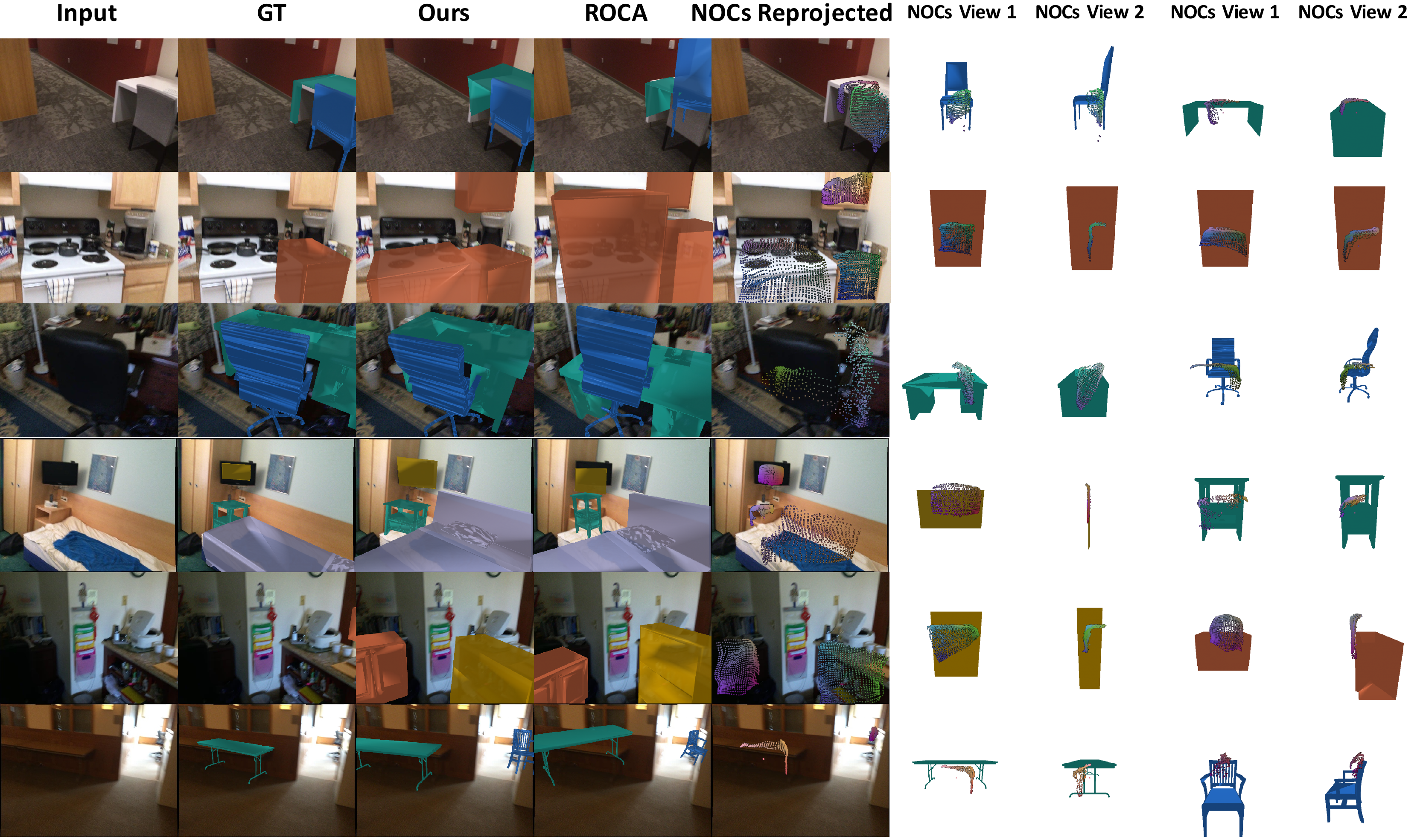}
    \vspace{-0.5cm}
    \caption{\textbf{Visualisation of ROCA's NOCs predictions.} For different inputs we show the GT CAD model alignment, our prediction and ROCA's prediction. Further we show the estimated NOCs overlayed in the canonical object frame from different views for relevant objects. We also show the NOCs reprojected back into the image under the predicted pose. One can observe that the predicted NOCs are often systematically offset from the actual 3D shape. Therefore even though the NOCs reprojected under the estimated pose roughly match the images the corresponding CAD alignments can be very wrong.  
}
    \label{roca_nocs}
\end{figure*}

ROCA \cite{roca} relies on predicting Normalised Object Coordinates (NOCs) for each pixel of the detected objects. NOCs are 3D object coordinates in a canonical frame in which objects have been aligned. However, here we demonstrate that aligning different object shapes with each other is not trivial. Figure \ref{issue_alignment} shows that even for two very similar shapes different alignments are possible depending on which object parts one wishes to align. This means that NOCs learned for one shape do not generalise well to other shapes. When attempting to predict NOCs we observe that ROCA \cite{roca} often predicts NOCs with a systematic offset (see Figure \ref{roca_nocs}). Here we show the NOCs predicted overlayed in the canonical object frame. One can see that the 3D coordinates predicted are often systematically above or below the actual object. This means that even though the reprojected NOCs roughly match the objects in the image the corresponding object alignment is very wrong.\\

% T ROCA:
% 20,833;10,0;22,845;8,962;11,538;40,714;14,136;17,699;11,392;17,569;24,226;

% R ROCA:
% 19,167;22,857;29,31;13,208;20,0;46,112;16,754;27,434;15,913;23,417;29,606;

% S ROCA:
% 1,667;24,286;22,845;6,604;12,692;43,916;17,277;21,239;9,584;17,79;24,93;

\section{Ablating our system with ROCA predictions}
\label{sec_ablating_roca_predictions}
The previous section showed that ambiguities in aligning different shapes with each other can lead to systematic offsets when predicting NOCs. A second issue with ROCA \cite{roca} is that object scale is directly regressed which is inaccurate and can produce very wrong scale estimates subsequently leading to poor translation estimates. We demonstrate both of these issues quantitatively by replacing either the scale, translation or rotation prediction of our system with ROCA's \cite{roca} prediction (see Table \ref{ablate_with_roca_preds}, lower half). The top half of the table is a copy of our main results table and is for reference only. Note that the row ``Ours + ROCA rot init'' uses the ROCA rotation predictions as an initialisation which is subsequently refined by our own predictions. In contrast the rows ``Ours + ROCA rotation/translation/scale'' show results when replacing the respective predictions with ROCA's predictions and keeping them fixed during the refinement process. We observe a noticeable drop in alignment accuracy when replacing our translation or scale predictions with ROCA's confirming the issues presented above. ROCA's rotation predictions are largely unaffected by inaccurate scale predictions and systematic offsets in NOCS and we observe that they perform similarly to ours.
\begin{table}[t]
    \centering
    \resizebox{\textwidth}{!}{
    \begin{tabular}{c|ccccccccc|cc}
    % {||p{3cm} | p{0.8cm} p{0.8cm} p{0.8cm} p{0.8cm} p{0.8cm} p{0.8cm} p{0.8cm} p{0.8cm} p{0.8cm} p{0.8cm}||} 
    
     Method & bathtub & bed & bin & bkshlf & cabinet & chair & display & sofa & table & \textbf{class} & \textbf{instance}\\ [0.5ex] 
     \hline
    % \toprule
     Number of Instances \# & 120 & 70 & 232 & 212 & 260 & 1093 & 191 & 113 & 553 & 9 & 2844 \\
     \hline
    % \midrule
     ROCA \cite{roca} & 20.8 & 8.6 & 26.3 & 9.0 & 13.1 & 39.9 & 24.6 & 10.6 & 12.7 & 18.4 & 25.0\\
    %  Ours & 13.3 & 31.4 & 30.6 & 11.8 & 18.1 & 46.7 & 22.5 & 18.6 & 17.5 & 23.4 & 30.0\\
    %  Ours + ROCA rot init & 25.0 & 25.7 & 38.8 & 14.2 & 20.4 & 53.0 & 25.1 & 23.9 & 18.6 & 27.2 & 34.4\\
    Ours & 25.8 & 25.7 & 24.6 & 14.2 & 20.8 & 51.5 & 17.8 & 28.3 & 15.4 & 24.9 & 31.8 \\
    Ours + ROCA rot init & 25.0 & 30.0 & 36.2 & 14.2 & 19.2 & 52.3 & 20.4 & 28.3 & 20.1 & 27.3 & 
    34.1\\
     \hline

     Ours + ROCA rotation & 19.2 & 22.9 & 29.3 & 13.2 & 20.0 & 46.1 & 16.8 & 27.4 & 15.9 & 23.4 & 29.6\\
     Ours + ROCA translation & 20.8 & 10.0 & 22.8 & 9.0 & 11.5 & 40.7 & 14.1 & 17.7 & 11.4 & 17.6 & 24.2
      \\
     Ours + ROCA scale & 1.7 & 24.3 & 22.8 & 6.6 & 12.7 & 43.9 & 17.3 & 21.2 & 9.6 & 17.8 & 24.9\\

    %  Ours + ROCA rotation & 20.0 & 24.3 & 30.6 & 12.7 & 18.1 & 46.1 & 21.5 & 21.2 & 15.6 & 23.3 & 29.6\\
    %  Ours + ROCA translation & 15.0 & 12.9 & 25.9 & 8.5 & 13.8 & 41.0 & 14.7 & 18.6 & 11.2 & 17.9 & 24.6
    %   \\
    %  Ours + ROCA scale & 2.5 & 20.0 & 22.8 & 6.6 & 11.2 & 41.6 & 16.2 & 18.6 & 8.3 & 16.4 & 23.4\\
    \end{tabular}}
    \caption{\textbf{Ablation of SPARC-Net with ROCA predictions.} The top half repeats the results presented in the main paper. The lower half shows results when we replace one of our predictions (rotation, translation or scale) with ROCAs \cite{roca} prediction. Note that those predictions are not refined with our own predictions. Row 3 ``Ours + ROCA rot init'' in contrast uses ROCA's rotation prediction as an initialisation which is subsequently refined with our own predictions.}
    \label{ablate_with_roca_preds}
\end{table}

\section{Limitations}
\label{sec_limitations}
This section lists limitations of the current approach which we plan to address in future works.\\
\textbf{Dense depth and surface normal predictions.}
Currently our method uses dense depth and surface normal estimates that are precomputed. If depth and surface normals are computed in realtime, predicting them sparsely only for relevant pixel coordinates may reduce inference time. Further for real applications our method could make use of additional sensory information such as LiDAR which provides naturally sparse depth inputs that can easily replace predicted depth values in our pipeline. Further for video applications event cameras \cite{event_camera} may be of particular interest as these are extremely fast and energy efficient as they only react to changes in light intensity, therefore providing sparse image data containing object edge information that is crucial for 3D shape estimation or CAD model alignment.\\
\textbf{Rotation predictions.}
Currently our handling of rotation is not elegant as at test time it requires four extra forward passes through our network to determine the initialisation of the rotation. This aspect could be improved by 
reprojecting the four different rotation initialisation simultaneously while at the same time predicting pose updates for all of them, but then only applying those pose updates to the rotation initialisation with the highest estimated probability.\\
\textbf{CAD model refinements.}
Currently, our approach is limited to refining an initial object pose, but not the initial object shape. In general all retrieval-based approaches for shape estimation are limited by the availability of a fitting CAD model. However, even with growing CAD model databases it is unrealistic that every object in the real world will have a precisely fitting CAD model in the database. Therefore it is important to deform a retrieved CAD model to better fit an observed object. This could be nicely achieved with the presented framework by in addition to 9 DoF pose updates predicting $N$-dim shape updates where $N$ are the numbers of parameters of some shape transformation. One possibiliy for such a shape transformation function are neural cages \cite{neural_cages}.\\
\textbf{Joint shape and pose predictions.}
Currently for reconstructing scenes every object is treated individually. However, this neglects important information contained within object-object relationships. These can contain information about the pose (e.g. two tables that are standing next to each other are likely to be aligned with each other) or the shape (e.g. chairs around a table are likely to have the same shape which can be a very important signal when dealing with strong occlusion.). Taking into account such information (e.g. by modelling it as a scene-graph \cite{scenecad} will further improve our shape and pose predictions.

\section{Visualisation Video}
\label{sec_vis_video}

\begin{figure*}[t]
    \centering
\includegraphics[width=1.0\linewidth]{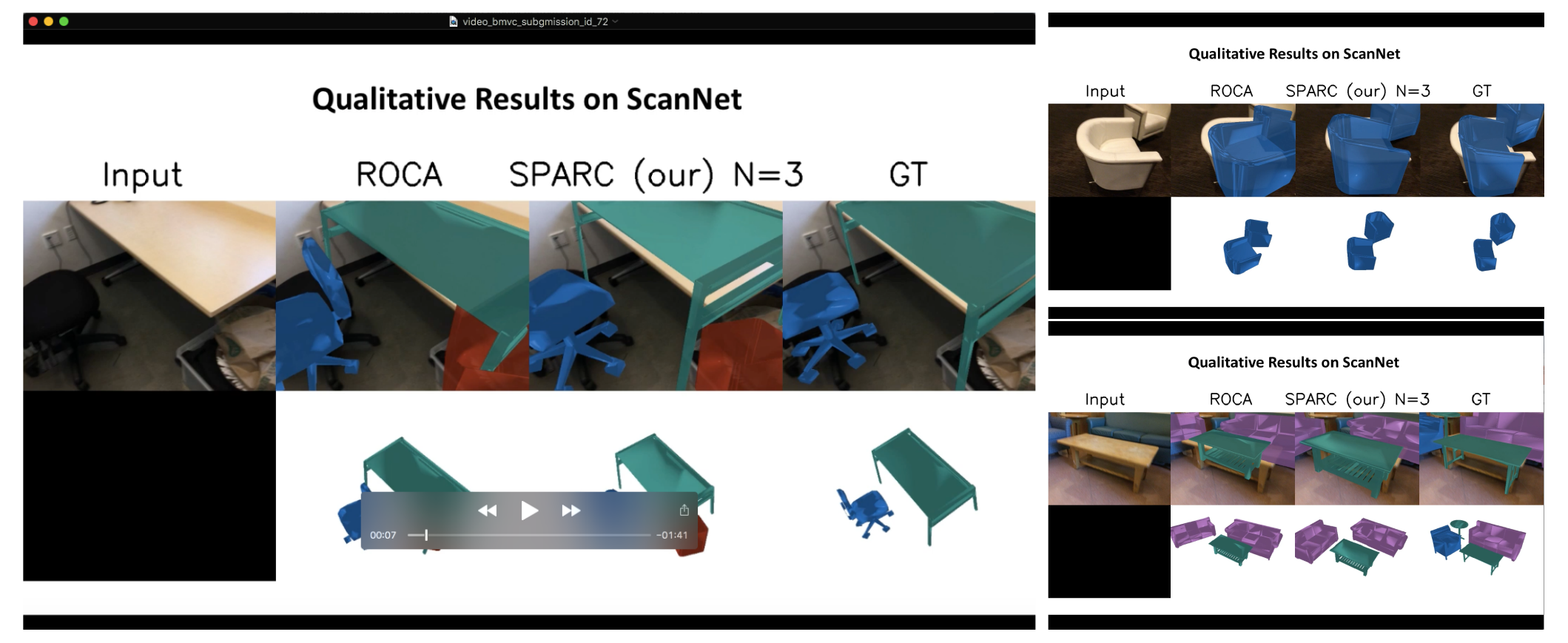}
    \caption{\textbf{Visualisation Video.} We visualise intermediate refinement steps on ScanNet \cite{scannet} in a video (\url{https://youtu.be/eVVW__0QGnM}).} 
    \label{vis_video}
\end{figure*}
In the visualisation video (\url{https://youtu.be/eVVW__0QGnM}) one can see nicely that our pose prediction network, SPARC-Net, is very robust to rough initialisation and able to significantly translate, rotate and scale CAD models to fit the objects observed in the image. The results demonstrate the advantage of an iterative procedure: the first refinement is usually a large pose update, transforming the often very bad initialisation to roughly match the pose of the object in the image. The second and third refinement in contrast are smaller pose updates that really align the CAD model with the objects in the image.

Note that for some images in the video the number of ground truth CAD models and the number of CAD models for which the pose is predicted do not correspond exactly, either due to missing 2D object detections or because of missing ground truth annotation as \cite{scan2cad} did not provide exhaustive CAD model annotation for ScanNet \cite{scannet}.
\end{document}